\documentclass[11pt]{article}

\usepackage{acl}
\usepackage{times}
\usepackage{latexsym}
\usepackage{amsmath}
\usepackage[T1]{fontenc}
\usepackage[utf8]{inputenc}
\usepackage{microtype}
\usepackage{multirow}
\usepackage{caption}
\usepackage{enumitem}
\usepackage{booktabs}   
\usepackage{graphicx}   
\usepackage[table]{xcolor}
\usepackage{colortbl}
\usepackage{amssymb}
\usepackage[ruled,vlined]{algorithm2e}  

\usepackage{tcolorbox}
\usepackage{todonotes}

\colorlet{VanillaColor}{yellow!5!white}
\colorlet{RAGColor}{green!5!white}
\colorlet{MemoryColor}{cyan!8!white}
\colorlet{StructuralMemoryColor}{purple!8!white}

\title{Chat2Workflow: A Benchmark for Generating Executable Visual Workflows with Natural Language}

\author{
  Yi Zhong${^{\spadesuit\heartsuit}}$\thanks{Equal contribution. \hspace{2em} $^\dagger$Corresponding Author.},
  Yijun Wang$^{\heartsuit}$\footnotemark[1],
  Buqiang Xu$^{\spadesuit}$\footnotemark[1],
  Zifei Shan$^{\heartsuit}$\footnotemark[1],
  Shuofei Qiao$^{\spadesuit}$, \\
  \textbf{Guozhou Zheng$^{\spadesuit}$},
  \textbf{Ningyu Zhang}$^{\spadesuit}$\footnotemark[2]\\
  $^\spadesuit$Zhejiang University ~$^\heartsuit$Tencent \\
  \texttt{\{zhongyi0212, zhangningyu\}@zju.edu.cn} \\
  \textbf{\url{https://github.com/zjunlp/Chat2Workflow}}
}

\begin{document}

\maketitle

\begin{abstract}
At present, executable visual workflows have emerged as a mainstream paradigm in real-world industrial deployments, offering strong reliability and controllability. However, in current practice, such workflows are almost entirely constructed through manual engineering: developers must carefully design workflows, write prompts for each step, and repeatedly revise the logic as requirements evolve---making development costly, time-consuming, and error-prone. To study whether large language models can automate this multi-round interaction process, we introduce \textbf{Chat2Workflow}, a benchmark for generating executable visual workflows directly from natural language, and propose \textbf{a robust agentic baseline} to improve performance. The benchmark is built from a large collection of real-world business workflows, with each instance designed so that the generated workflow can be transformed and directly deployed to practical workflow platforms such as Dify and Coze. Experimental results show that while state-of-the-art language models can often capture high-level intent, they struggle to generate correct, stable, and executable workflows, especially given complex and evolving requirements. Although our agentic baseline yields up to 6.05\% resolve rate gains, the remaining real-world gap positions Chat2Workflow as a foundation for advancing industrial-grade automation.
\end{abstract}


\section{Introduction}

Large language models now form the core of many real-world applications, increasingly deployed through agent-based systems \cite{zhao2023survey,wang2024survey,xi2025rise}. 
Existing systems can be broadly divided into ReAct agents  \cite{yao2022react} and agentic workflows \cite{zeng2023flowmind,fan2024workflowllm,qiaobenchmarking}. 
While ReAct agents are flexible, prior work and industry experience suggest that agentic workflows are better suited for reliable and controllable industrial use \cite{shi2025flowagent}. 
Recent interviews \cite{pan2025measuring} show that simple methods dominate production systems: more than \textbf{70\%} of real-world agent deployments rely on off-the-shelf language models without any weight tuning, instead \textbf{using explicit workflows for orchestration}. 
This trend is reflected in widely used platforms such as Dify\footnote{\url{https://dify.ai/}} and Coze\footnote{\url{https://www.coze.com/}}, where agent behavior is primarily predefined through workflows.

\begin{figure}[t]
    \centering
    \includegraphics[width=1.0\linewidth, trim=20 15 20 5mm, clip]{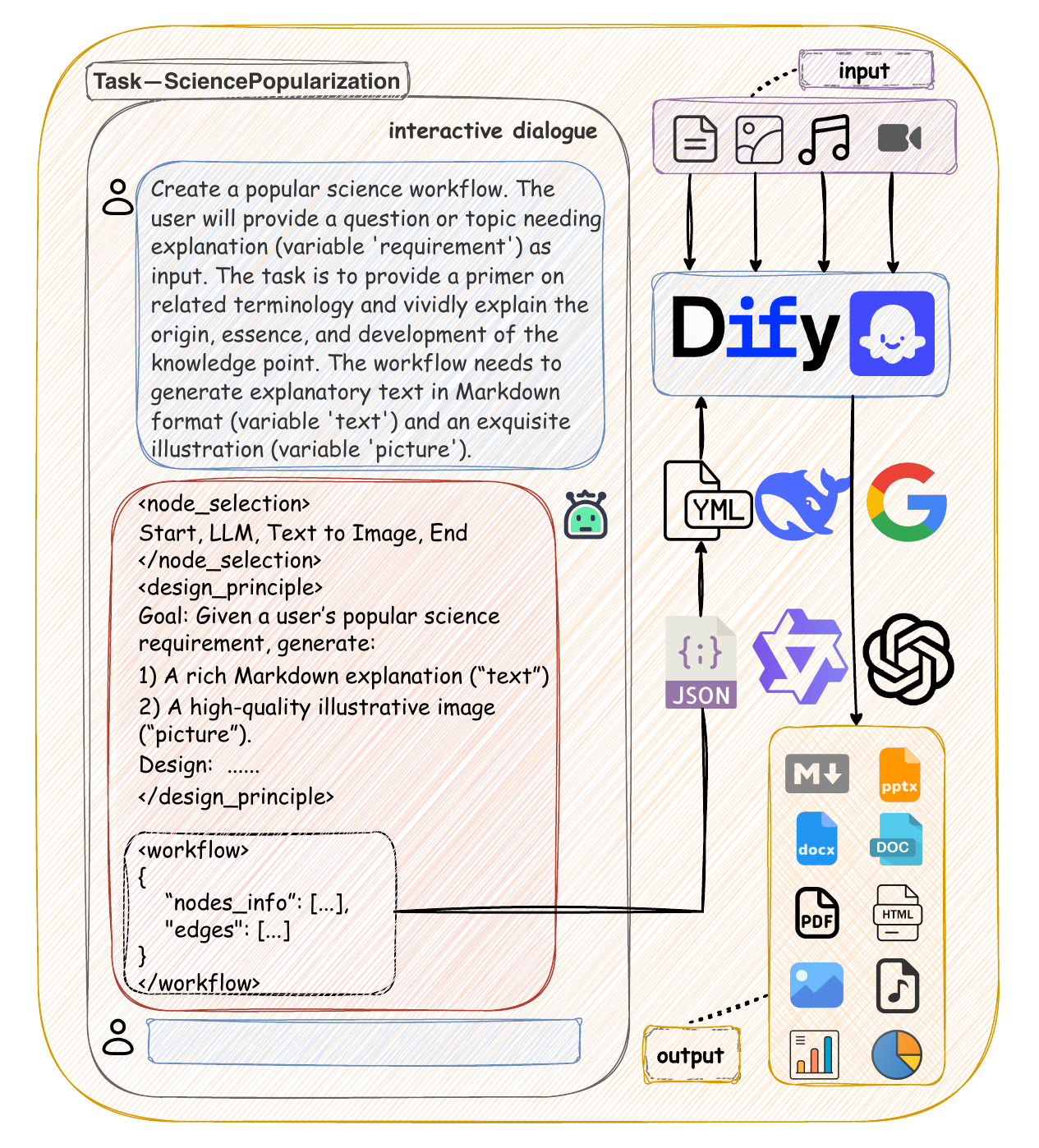}
    \vspace{-2mm}
    \caption{\textbf{An example task in Chat2Workflow}, which features realistic, variable natural-language instruction inputs and produces outputs that can be directly transformed and integrated into real-world workflow platforms (e.g., Dify and Coze).}
    \label{fig:intro_picture}
\end{figure}

Despite their prevalence, agentic workflows are still largely created by manual engineering. 
Human developers must translate natural language requirements into structured workflows, which is time-consuming and error-prone. 
This raises a natural question: \textbf{can agentic workflows be automatically generated from natural language?}
Achieving this is challenging for two reasons. 
First, real-world requirements are often complex and implicit, making it difficult to infer correct control flow and tool usage from natural language alone. 
Second, user requirements frequently change, requiring workflows to be revised or regenerated while preserving correctness and consistency.

\begin{figure}[t]
    \centering
    \includegraphics[width=0.6\linewidth]{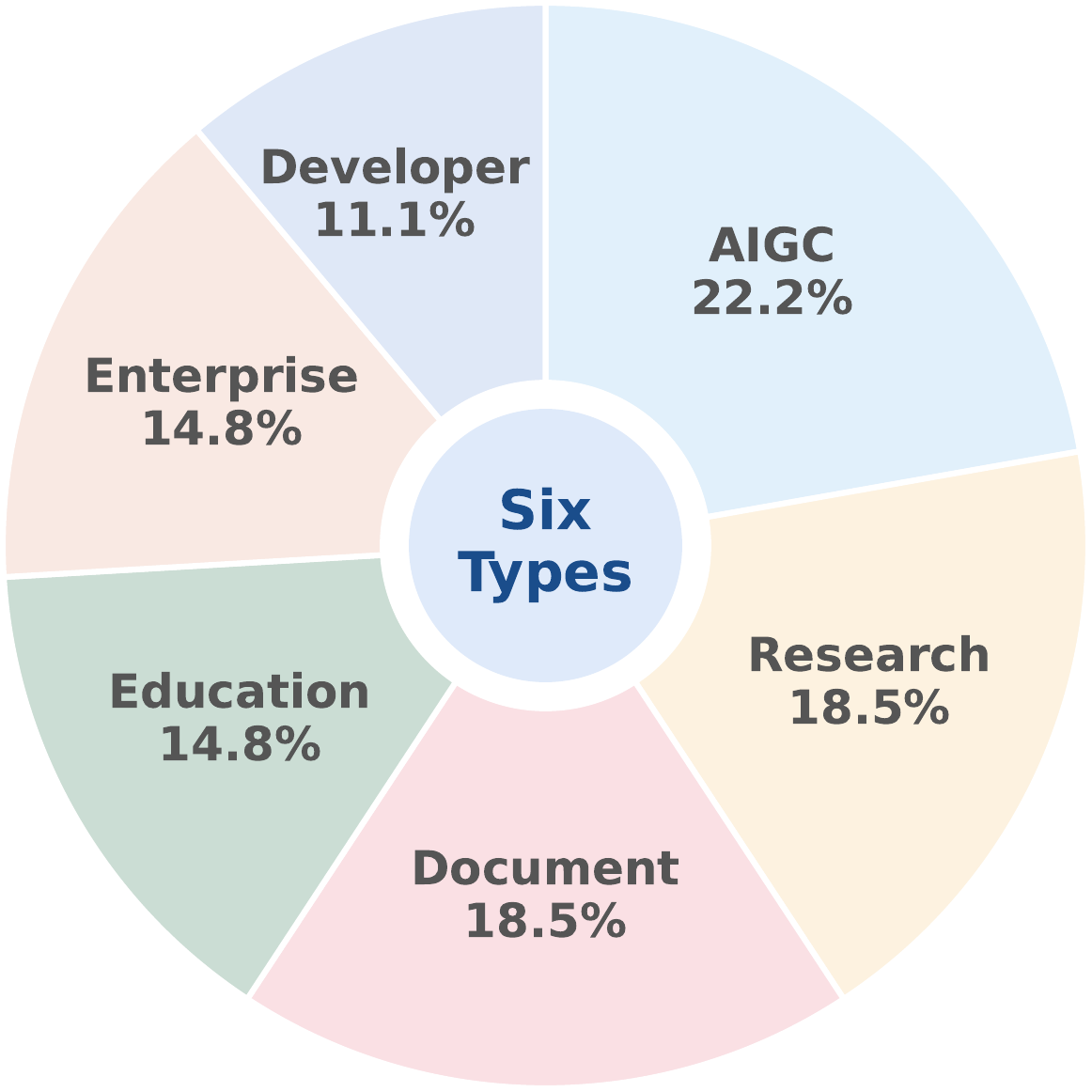}
    \vspace{-2mm}
    \caption{Distribution of task types in Chat2Workflow. The benchmark covers six domains: AIGC, Research, Document, Education, Enterprise, and Developer.}
    \label{fig:task_distribution}
\end{figure}

To systematically study this, we introduce \textbf{Chat2Workflow} (MIT license), the first benchmark focused on generating executable visual workflows from natural language (Figure~\ref{fig:intro_picture}). 
It comprises 237 human-annotated instances across six domains (Figure~\ref{fig:task_distribution}) to evaluate whether workflows meet the expected requirements for execution. Each round is equipped with realistic user requests and target workflow components across diverse domains and complexity levels.

The benchmark evaluates a model's ability to infer structures, select tools, and produce executable, intent-aligned workflows. 
Experiments show that current state-of-the-art language models can often capture high-level intent but struggle to generate correct and stable workflows, especially as tasks become more complex or requirements evolve.
To address this, we propose an error-driven agentic baseline with auto-repair mechanisms. 
Although it improves absolute resolve rates by 6.05\%, the remaining performance gap highlights executable workflow generation as a key automation bottleneck, positioning Chat2Workflow as a crucial benchmark for future research in structured reasoning and adaptive workflow synthesis.





\SetKw{KwNotation}{Notation:}
\section{Constructing Benchmark}
\subsection{Workflow Orchestration Platforms}
\textbf{Overview} Dify and Coze are mainstream platforms for building agentic workflows by connecting functional nodes. These workflows are visualized as Directed Acyclic Graphs on the frontend and stored as structured YAML files on the backend. The platforms execute tasks node-by-node, automatically handling tool invocations, inference, and data flow. Taking Dify as a representative example, it provides essential built-in nodes (e.g., Code, If-Else) alongside hundreds of community-driven extensions (e.g., Google Search), enabling the resolution of highly diverse and complex tasks.



\begin{figure*}[t]
    \centering
    \includegraphics[width=1.0\linewidth, trim=15 15 5 5mm, clip]{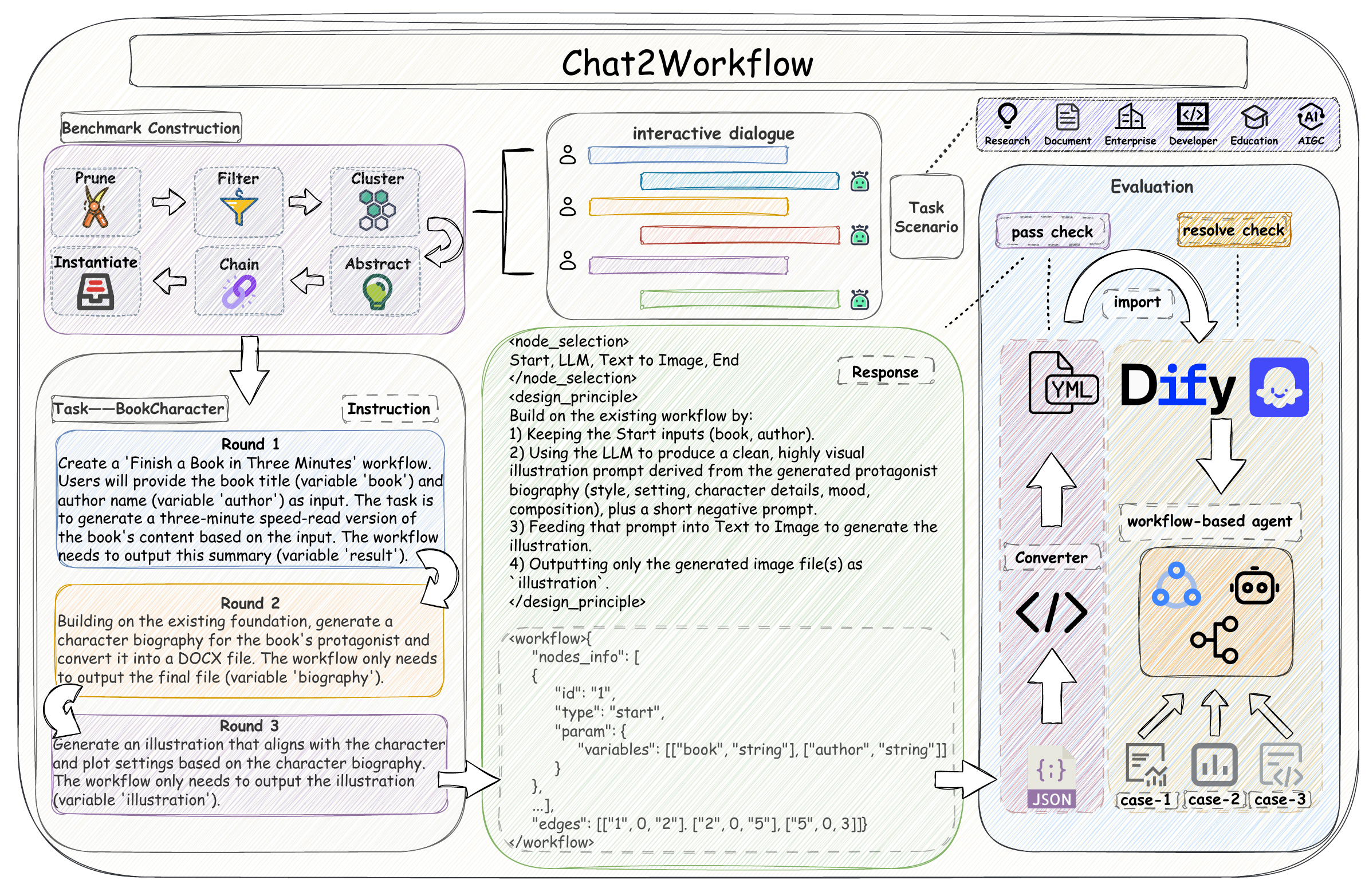}
    \vspace{-2mm}
    \caption{Overview of Chat2Workflow benchmark construction and evaluation framework. \textbf{Left}: The process of constructing multi-turn instructions involves six sequential stages: Prune, Filter, Cluster, Abstract, Chain, and Instantiate. \textbf{Center}: Users interact with LLMs through dialogue, and the model generates workflows in JSON format with CoT reasoning (node selection, design principles, and structured workflow). \textbf{Right}: The workflows span six distinct task scenarios. The JSON workflow is converted to YAML format, uploaded to the platforms for execution, and evaluated against test cases to compute Pass Rate and Resolve Rate.}
    \label{fig:main_pic}
\end{figure*}


\subsection{Task Formulation}
We formalize the generation task of the workflow as a multi-turn interaction process.
Given a task instruction for the current round and historical interaction dialogues, we aim to enable language agents to generate graph-structured workflows represented in JSON, which describes the nodes and connectivity in a structured manner. 
Formally, given the instruction $q$, historical dialogues $H$ and language agents $\mathcal{M}_\theta$, the workflow generation can be represented as:
\begin{equation}
    \mathcal{G}(\mathcal{V},\mathcal{E}) \leftarrow \mathcal{M_\theta}(q, H),
\end{equation}
where $\mathcal{G}$ is a Directed Acyclic Graph with node list $ \mathcal{V} = \{v_1,v_2, ..., v_{|\mathcal{V}|}\} $ comprising nodes or tools with predefined parameters populated and edges $ \mathcal{E} = \{(v_i, p ,v_j)\}, 1 \le i \neq j \le n, p \in \mathbb{Z}_{\ge 0}$ indicating that the source node $v_i$ is connected to the target node $v_j$ through the output branch port $p$. 

Directly generating executable workflows is challenging for LLMs. To address this, we use a Chain-of-Thought (CoT) approach that outputs a specially designed simplified JSON. The model generates selected nodes $\hat{\mathcal{V}}$, planning rationale $p$, and the JSON representation $r$, wrapped in <node\_selection>, <design\_principle> and <workflow> tags. This JSON is then converted into an executable YAML file. It can be formalized as:
\begin{equation}
    \mathcal{G}(\mathcal{V},\mathcal{E}) \leftarrow r \leftarrow \{\hat{\mathcal{V}},p,r\} \leftarrow \mathcal{M_\theta}(q, H),
\end{equation}
where $\mathcal{V}$ and $\mathcal{E}$ correspond to the \texttt{nodes\_info} and \texttt{edges} fields in JSON respectively.


\subsection{Node Selection and Processing}
We selected the 20 most frequently used nodes to capture major real-world scenarios.
To reduce generation complexity, we simplified the nodes' input/output (I/O) variable interfaces: the language agent only needs to provide key-value pairs for essential primary variables, while secondary variables default.
In order to endow the language agent with the usage of each node and corresponding I/O interfaces, we  craft node knowledge base for 20 node types, serving as a crucial part of the system prompt used for understanding the node ecosystem. 
Based on these principles, we extract and rewrite essential information from official node documents into a predefined structure comprising <type>, <description>, <parameter>, <referable\_variables>, and <supplementary\_information>.





\subsection{Dataset Construction}

Figure \ref{fig:main_pic} outlines our construction and evaluation framework. To build a multi-turn workflow dataset, \textbf{three independent CS undergraduate volunteers} processed production-level Dify and Coze configurations collected from official sources and GitHub through a systematic pipeline:

\noindent \textbf{Pruning and Filtering.} We first substituted unsupported workflow nodes with functional equivalents or pruned unviable functional components. Workflows with poor testability were discarded.

\noindent \textbf{Clustering and Abstracting.} The remaining workflows were clustered into six task scenarios: Research, Document, Enterprise, Developer, Education, and AIGC (details in Appendix \ref{data:appendix}). Within each cluster, workflows were abstracted into natural language representations of their core functions, which are closely tied to essential key nodes. 

\noindent \textbf{Chaining.} To construct multi-turn tasks (2–4 rounds), volunteers manually chained related workflows, treating each as a single turn. Each round's instruction sequentially adds, modifies, or deletes functions from the previous round while maintaining logical consistency. We maximized variation across graph topologies, I/O types, and file formats, strictly distinguishing between function modifications and complete overwrites. I/O variables are explicitly formatted in parentheses, and unavoidable key nodes are recorded in advance as hints.

\noindent \textbf{Instantiating.} To enable end-to-end evaluation, we instantiated each instruction with three test cases that comprehensively cover historical requirements and multi-branch logic. Test cases were sourced hierarchically: (1) relevant Hugging Face datasets adapted to our I/O interfaces, (2) real-world internet materials, and (3) LLM-synthesized data. Reference answers are provided for deterministic outputs, whereas open-ended general cases consist solely of the provided inputs.

\subsection{Quality Control and Dataset Statistics}


We implement a strict \textbf{cross-validation review process}, with \textbf{three volunteers} auditing each other's data to ensure correctness, completeness and privacy.
To pass, the instructions must allow the target workflow's key nodes and core functions to be manually recreated without ambiguity, exactly matching the creator's intent.
Cases with invalid intermediate steps, time-varying sources, or incoherent behaviors are systematically removed.

Following this rigorous validation, we curated Chat2Workflow, a high-quality dataset consisting of 27 tasks with 79 multi-turn instructions. Accompanied by 237 test instances (three per instruction), it robustly assesses whether generated workflows adhere to complex execution requirements.

\subsection{Evaluation Protocol}


Unlike standard text or image generation, workflows must be evaluated not only on their representational format but also on their executability and outcomes. Thus, we adopt a two-stage progressive evaluation: \textit{pass rate} assesses structural and logical validity, while \textit{resolve rate} evaluates functional execution correctness.

\begin{algorithm}[h]
\small
\caption{Pass Rate Pipeline}
\label{alg:pass_rate}
\KwData{Model response $\mathcal{R}$, Reference variables $\mathcal{V}_{ref}$, Key nodes $N_{key}$}
\KwNotation{\textnormal{Selected nodes} $\hat{\mathcal{V}}$, \textnormal{Planning rationale} $p$, \textnormal{JSON file} $r$, \textnormal{YAML file} ${F_{yaml}}$, \textnormal{Workflow} $\mathcal{G}$}

\tcp{Step 1: Format Check}
$\hat{\mathcal{V}}, p, r \leftarrow \textsc{Parse}(\mathcal{R})$\;

\If{$\mathcal{R}$ lacks required tags \textbf{or} JSON $r$ not parsable}{
    \Return $False$\;
}

\tcp{Step 2: Conversion \& Import}

${F_{yaml}} \leftarrow \textsc{JsonToYaml}(r)$\;
\If{conversion fails}{\Return $False$\;}
$\mathcal{G} \leftarrow \textsc{Import}(F_{yaml})$\; 

\tcp{Step 3: Variable Consistency}
\If{$\textsc{ExtractVars}(r) \neq \mathcal{V}_{ref}$}{\Return $False$\;}

\tcp{Step 4: Logical Validity}
\If{$f_{LLM}(\hat{\mathcal{V}}, p, r)$ indicates inconsistency} {\Return $False$\;}
\If{$f_{LLM}(N_{key},\hat{\mathcal{V}})$ indicates $N_{key} \not\subseteq \hat{\mathcal{V}}$} {\Return $False$\;}

\Return $True$\;
\end{algorithm}



\textbf{Pass rate} measures the proportion of subtasks that successfully pass the sequential checks in Algorithm~\ref{alg:pass_rate}. Specifically, a subtask passes if its JSON representation aligns with the CoT sequence and predefined variables, and converts into an importable YAML file. It is defined as:
\begin{equation}
    \text{\%Pas.} = \frac{N_{\text{success\_subtask}}}{N_{\text{total\_subtask}}},
\end{equation}
where $N_{\text{success\_subtask}}$ and $N_{\text{total\_subtask}}$ denote the number of passing subtasks and the total number of subtasks, respectively.







\textbf{Resolve rate} measures the proportion of test cases where the generated workflow correctly executes and produces outputs satisfying all instruction requirements in Algorithm~\ref{alg:resolve_rate}. It is defined as:
\begin{equation}
    \text{\%Res.} = \frac{N_{\text{success\_case}}}{N_{\text{total\_case}}},
\end{equation}
where $N_{\text{success\_case}}$ and $N_{\text{total\_case}}$ represent the number of successful test cases and the total number of test cases, respectively.

\begin{algorithm}[h]
\small
\caption{Resolve Rate Pipeline}
\label{alg:resolve_rate}
\KwData{Workflow $\mathcal{G}$, instructions $\mathcal{I}$, Test case $\mathcal{T} = \{input, ref\_output\}$}
\KwNotation{\textnormal{Execution output} $\mathcal{O}$}

\tcp{Step 1: Execution Check}
$\mathcal{O} \leftarrow \textsc{Execute}(\mathcal{G}, \mathcal{T}.\text{input})$\;
\If{execution fails with errors \textbf{or} $\mathcal{O} = \emptyset$}{\Return $False$\;}

\tcp{Step 2: Output Validation}
$\mathcal{O}_{file}, \mathcal{O}_{text} \leftarrow \textsc{SplitOutput}(\mathcal{O})$\;
\If{$\mathcal{O}_{file} \neq \emptyset$ \textbf{and} file types/extensions mismatch}{\Return $False$\;}
\If{$\mathcal{O}_{text} = \emptyset$}{\Return $True$\;}
\If{$f_{LLM}(\mathcal{I},\mathcal{O}_{text}, \mathcal{T}.\text{input}, \mathcal{T}.\text{ref\_output})$ confirms all requirements met}{\Return $True$\;}

\Return $False$\;
\end{algorithm}

\definecolor{closedbg}{RGB}{204, 153, 204} 
\definecolor{openbg}{RGB}{165, 205, 238}

\begin{table*}[t]
\centering
\resizebox{\linewidth}{!}{%
\begin{tabular}{l | cc | cc | cc | cc | cc | cc | cc}
\toprule[1.5pt]

\multirow{2}{*}{\textbf{Model}} & 
\multicolumn{2}{c|}{\textbf{Research}} & 
\multicolumn{2}{c|}{\textbf{Document}} & 
\multicolumn{2}{c|}{\textbf{Enterprise}} & 
\multicolumn{2}{c|}{\textbf{Developer}} & 
\multicolumn{2}{c|}{\textbf{Education}} & 
\multicolumn{2}{c|}{\textbf{AIGC}} & 
\multicolumn{2}{c}{\textbf{Average}} \\
\cmidrule(lr){2-3} \cmidrule(lr){4-5} \cmidrule(lr){6-7} \cmidrule(lr){8-9} \cmidrule(lr){10-11} \cmidrule(lr){12-13} \cmidrule(lr){14-15}
 & \text{\%Pas.} & \text{\%Res.} & \text{\%Pas.} & \text{\%Res.} & \text{\%Pas.} & \text{\%Res.} & \text{\%Pas.} & \text{\%Res.} & \text{\%Pas.} & \text{\%Res.} & \text{\%Pas.} & \text{\%Res.} & \text{\%Pas.} & \text{\%Res.}  \\
\midrule

\rowcolor{closedbg}
\multicolumn{15}{c}{\textbf{\textit{Closed-Sourced}}} \\
\midrule
GPT-5.1  & 48.89 & 41.48 & 53.33 & 40.74 & 50.00 & 33.33 & 20.83 & 8.33 & 36.36 & 19.19 & 57.41 & 45.06 & 47.26 & 34.46 \\
GPT-5.2     & 68.89 & 53.33 & 66.67 & 50.37 & \textbf{86.11} & 58.33 & 70.83 & 38.89 & 75.76 & 54.55 & 48.15 & 32.10 & 67.51 & 47.40 \\
Claude-Sonnet-4.5      & 68.89 & 56.30 & 75.56 & 62.22 & 61.11 & 27.78 & \textbf{75.00} & 31.94 & \textbf{90.91} & \textbf{60.61} & 62.97 & 41.97 & 71.31 & 47.96 \\
Gemini-3-Pro-Preview & \textbf{77.78} & \textbf{62.96} & \textbf{82.22} & \textbf{69.63} & \textbf{86.11} & \textbf{64.82} & 70.83 & \textbf{44.44} & \textbf{90.91} & 59.60 & \textbf{74.08} & \textbf{54.32} & \textbf{80.17} & \textbf{60.20}\\

\midrule

\rowcolor{openbg}
\multicolumn{15}{c}{\textbf{\textit{Open-Sourced}}} \\

\midrule
Qwen-3-8B & 40.00 & 21.48 & 8.89 & 4.45 & 8.33 & 0.00 & 12.5 & 1.39 & 3.03 & 0.00 & 7.41 & 6.79 & 13.92 & 6.61 \\
Qwen-3-14B & 26.67 & 18.52 & 33.33 & 14.08 & 13.89 & 0.00 & 20.83 & 11.11 & 27.27 & 0.00 & 35.18 & 22.22 & 27.43 & 12.38 \\
Qwen-3-32B & 31.11 & 22.22 & 28.89 & 13.33 & \textbf{30.56} & \textbf{5.55} & 50.00 & 15.28 & 33.33 & 23.23 & \textbf{48.15} & \textbf{29.63} & 36.71 & 19.13 \\
Qwen-3-235B-A22B & 35.55 & \textbf{32.59} & 51.11 & \textbf{34.07} & 19.45 & 1.85 & \textbf{62.50} & \textbf{20.83} & 39.39 & \textbf{29.29} & 42.59 & 23.46 & 40.93 & \textbf{24.47} \\
Qwen-3-Coder-480B-A35B-Instruct & \textbf{53.33} & 29.63 & \textbf{57.78} & 28.89 & 25.00 & 1.85 & 25.00 & 0.00 & \textbf{45.45} & 17.17 & 37.03 & 23.46 & \textbf{42.19} & 19.13 \\

\midrule
GLM-4.6    & 62.22 & \textbf{57.04} & 60.00 & 39.26 & 63.89 & 24.07 & \textbf{45.83} & 5.55 & 72.73 & 25.25 & 64.82 & 39.51 & 62.45 & 35.02 \\
GLM-4.7    & \textbf{80.00} & 54.82 & \textbf{62.22} & \textbf{45.18} & \textbf{72.22} & \textbf{44.45} & 41.67 & \textbf{19.45} & \textbf{78.79} & \textbf{52.53} & \textbf{70.37} & \textbf{50.61} & \textbf{69.20} & \textbf{46.55}\\

\midrule
Deepseek-V3.1    & \textbf{51.11} & \textbf{42.22} & \textbf{51.11} & 34.07 & \textbf{61.11} & \textbf{24.07} & \textbf{20.83} & 2.78 & \textbf{45.45} & \textbf{32.32} & \textbf{62.96} & \textbf{40.12} & \textbf{51.48} & \textbf{32.07} \\
Deepseek-V3.2    & \textbf{51.11}  & 40.74 & 46.67 & \textbf{36.29} & 27.78 & 1.85 & 16.67 & \textbf{4.17} & 42.42 & 22.22 & 57.41 & 39.51 & 43.46 & 27.43 \\

\midrule
Kimi-K2-Instruct & 51.11 & 40.74 & 37.78 & 27.40 & 11.11 & 8.33 & 29.17 & 5.56 & 21.21 & 13.13 & 51.85 & 35.19 & 36.29 & 24.61 \\
Kimi-K2-Thinking & \textbf{55.56} & \textbf{44.44} & \textbf{66.67} & \textbf{46.67} & \textbf{52.78} & \textbf{25.93} & \textbf{62.50} & \textbf{26.39} & \textbf{27.27} & \textbf{15.15} & \textbf{72.22} & \textbf{43.83} & \textbf{57.81} & \textbf{36.00}\\
\bottomrule[1.5pt]
\end{tabular}
}
\vspace{-2mm}
\caption{Experimental results across six task domains for 4 closed-source and 11 open-source models. Pass Rate (\%Pas.) measures format correctness, while Resolve Rate (\%Res.) measures actual problem-solving capability. The best results within each model series are marked in \textbf{bold}.}
\label{tab:reproduced_structure}
\end{table*}

We employ GPT-5.1 for evaluation, sampling 355 instances for the pass stage and 463 for the resolve stage. To ensure reliability, \textbf{three independent volunteers} reviewed the LLM judge's reasoning and verdicts. Discrepancies were handled by requiring strict unanimity: the agreement rate is defined as the percentage of cases where all three evaluators independently confirmed that the LLM strictly followed the evaluation prompt's constraints (format, content, logic, and instruction following). Under this strict protocol, the LLM achieved 97.18\% agreement for the pass stage and 93.09\% for the resolve stage.
Cross-model validation with DeepSeek-V3 demonstrates consistent stability (96.62\% and 91.79\%). However, it prioritizes formal compliance over deep semantic accuracy, unlike the stringent GPT-5.1.
The exact prompts are detailed in Appendix (Figures~\ref{prompt:pass rate} and~\ref{prompt:resolve rate}).




\begin{figure*}[t]
    \centering
    \includegraphics[width=1.0\linewidth, trim=5 10 10 5mm, clip]{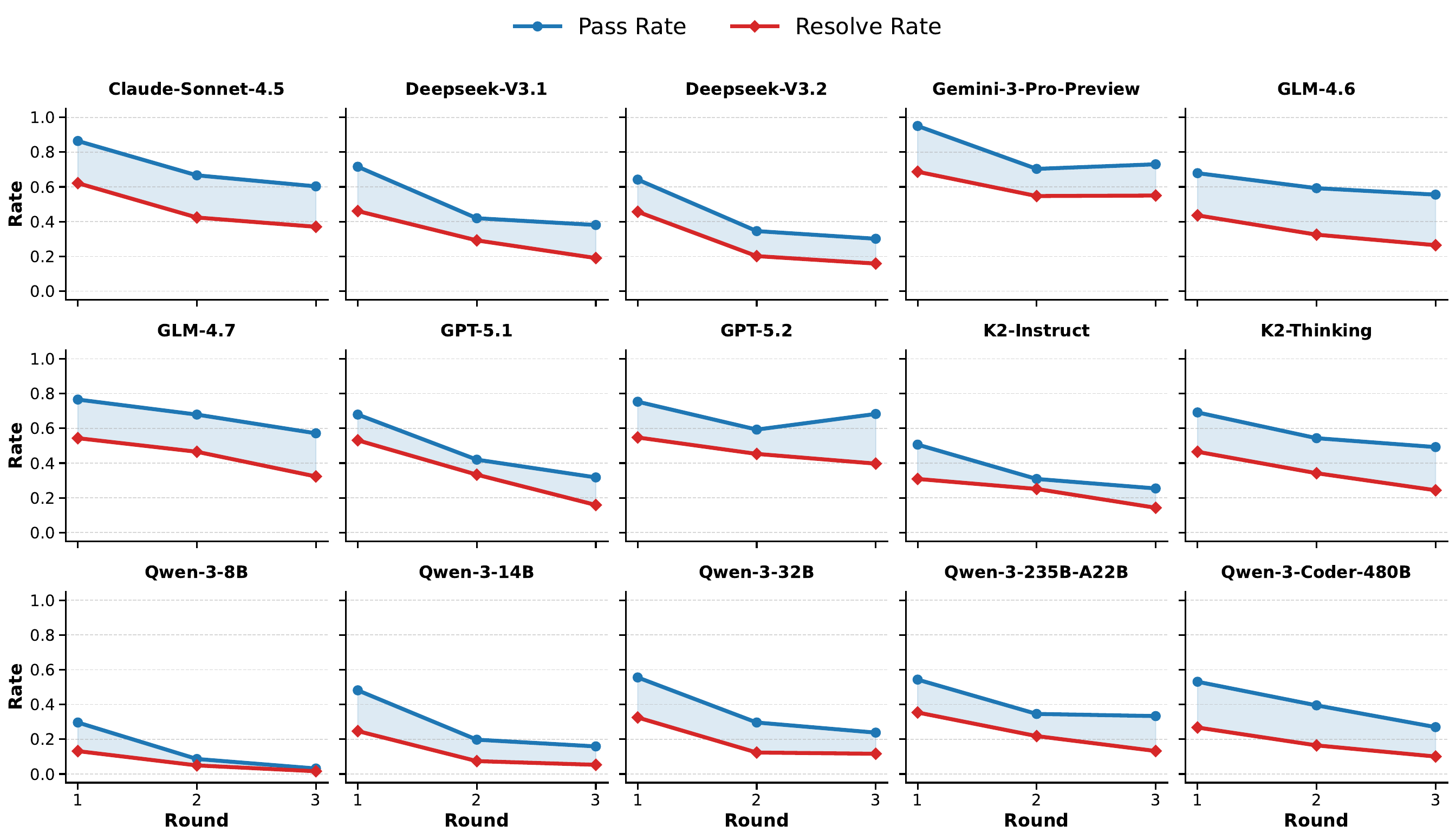}
    \vspace{-2mm}
    \caption{Performance degradation across dialogue rounds. We show the Pass Rate and Resolve Rate for all 15 models across the first three dialogue rounds. Most models exhibit a steady decline in both metrics as the number of interaction rounds increases, indicating the challenge of maintaining workflow quality under evolving requirements.}
    \label{fig:overview}
\end{figure*}

\section{Experiments}
\subsection{Experimental Setup}


To thoroughly assess workflow generation capabilities, we validate 15 representative models on the Chat2Workflow benchmark, reporting the average results across \textbf{three independent runs}: 



1) Four \textbf{closed-source} LMs: GPT-\{5.1, 5.2\}, Claude-Sonnet-4.5, and Gemini-3-Pro-Preview. 

2) Eleven \textbf{open-source} LMs: Qwen-3-\{8B, 14B, 32B\}, Qwen3-235B-A22B, Qwen3-Coder-480B-A35B-Instruct, GLM-\{4.6,4.7\}, DeepSeek-\{V3.1,V3.2\}, Kimi-K2-\{Instruct, Thinking\}.

Furthermore, to explore the performance upper bound, we select two representative models (GPT-5.1 and GPT-5.2) for an advanced agentic evaluation, which is detailed in Section 3.5.

The experiments are conducted on the 1.9.2 version of Dify. More details are in Appendix \ref{app:exp}.

\subsection{Main Results}
Table \ref{tab:reproduced_structure} presents our experimental results, which we use to analyze the following research questions.


\textbf{Q1: What is the gap between the workflow pass rate and resolve rate?}
For each task, we report both the pass rate and resolve rate to evaluate format compliance and practical usability, respectively. All models exhibit a lower resolve rate than pass rate. GLM-4.6 shows the largest disparity, averaging 27.43\% and surging to 47.48\% in the Education scenario. Although Kimi-K2-Instruct maintains a narrow gap of 11.68\%, its absolute performance remains poor compared to the highest resolve rate (60.20\%), particularly dropping in the Developer scenario. These results demonstrate that a syntactically legal workflow does not guarantee successful execution. Pass rates are necessary but insufficient, often inflating performance evaluations.


\textbf{Q2: To what extent do existing LLM agents approach the capabilities of real workflow design experts?}
Our benchmark largely represents real-world scenarios through its use of typical workflows, interactive creation mechanisms, and file I/O requirements. Evaluated purely on end-to-end task resolution, Gemini-3-Pro-Preview performs best with an absolute resolve rate of 60.20\%. However, It remains far below the threshold for a reliable, deployable planner. Even the best open-source model, GLM-4.7, achieves only a 46.55\% resolve rate and shows significant weakness in Developer tasks. Given our limited set of node types, real-world workflow creation is inherently more complex. Consequently, we anticipate that even leading models will struggle further when confronted with larger node vocabularies.

\textbf{Q3: How do scale, alignment, and reasoning impact workflow generation performance?} 
1) \textbf{Scale.} Increasing model parameters directly enhances performance. The Qwen-3 series demonstrates stable and significant improvements as its size scales from 8B to 32B and ultimately to 235B.
2) \textbf{Alignment.} While post-training improves format compliance, it does not guarantee problem-solving efficacy. For instance, in the Developer scenario, GLM-4.6 achieves a higher pass rate than GLM-4.7 (45.83\% vs. 41.67\%) but a drastically lower resolve rate (5.55\% vs. 19.45\%). Thus, optimizing solely for formatting may not effectively boost actual workflow usability.
3) \textbf{Reasoning.} Models equipped with reasoning capabilities consistently outperform standard instruction-tuned variants. Despite its massive scale, Qwen-3-Coder-480B-A35B-Instruct underperforms the smaller Qwen-3-235B-A22B in resolve rate (19.13\% vs. 24.47\%). Similarly, Kimi-K2-Thinking surpasses its instruct counterpart across all task domains, proving that explicit reasoning mechanisms are crucial for effective and robust workflow generation.

\begin{figure}[t]
    \centering
    \includegraphics[width=1.0\linewidth, trim=10 15 10 0mm, clip]{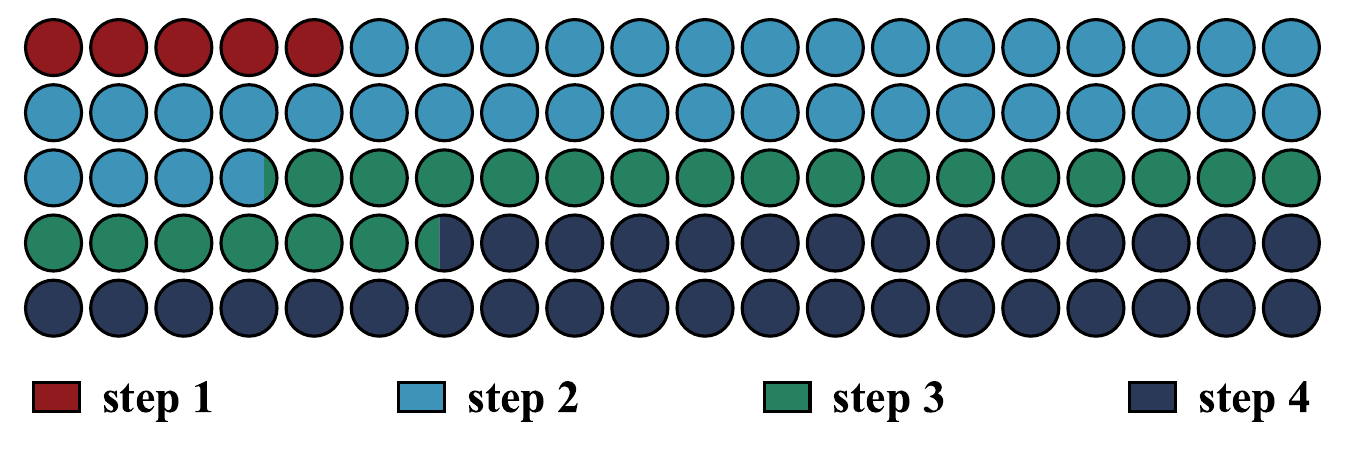}
    \vspace{-2mm}
    \caption{Error probabilities across the four steps of the pass stage (corresponding to Algorithm \ref{alg:pass_rate}).}
    \label{fig:error_distribution}
\end{figure}

\textbf{Q4: What shall we do to enhance the workflow generation capability of LLM agents?}
Analyzing failures from both the pass stage (see Figure \ref{fig:error_distribution} for the error distribution) and the resolve stage, we categorize the underlying errors into four primary dimensions. 1) \textbf{Format.} Syntactic and parsing rule violations, such as unclosed JSON brackets, redundant code fences, and improper character escaping. 2) \textbf{Graph.} Topological and structural flaws, including forward references violating topological sorting, hallucinated node connections, isolated nodes, and invalid cross-boundary edges connecting the inner and outer canvases of an \texttt{iteration} node. 3) \textbf{Semantics.} Cognitive misalignment regarding operational rules and node functionalities, manifested as failures to adhere to specific node configurations, violations of environment constraints, incorrect node types, or incompatible variable references. 4) \textbf{Consistency.} Functional or logical contradictions within the generation. This includes misalignments between the workflow's actual function and the user's core intent, mismatches with requested I/O variables, and logical discrepancies among different components of the model response or between consecutive turns.
Therefore, addressing these critical bottlenecks is key to advancing LLM agents' workflow generation capabilities.

\subsection{Analysis of User Requirement Changes from Chat Interactions}
Each task consists of two to four dialogue rounds. In each round, workflow requirements are iteratively modified based on historical context, presenting a significant challenge to the models' long-horizon instruction following.

To ensure robust representations, we analyze the evaluation metrics over the first three rounds (Figure~\ref{fig:overview}), omitting the fourth round due to its limited sample size. For most models, workflow quality consistently declines as interactions increase, corroborating the difficulty of handling cumulative modifications. 
The slight improvement observed for GPT-5.2 in the third round merely results from a changing denominator, as two-round tasks drop out of the calculation scope.


Furthermore, the degradation curve flattens over time. Rather than degrading at a uniform pace, the decline diminishes in later rounds. This indicates that workflows surviving the initial modifications act as a robust baseline. Once a model grasps the core logic in early turns, its capacity to accommodate further incremental changes stabilizes, reflecting a shift from drastic qualitative failures to minor quantitative errors.

\begin{figure*}[t]
    \centering
    \includegraphics[width=1.0\linewidth, trim=30 30 60 50mm, clip]{figure/bad_case_study.pdf}
    \vspace{-2mm}
    \caption{Bad case analysis for the StudyPlanner task. We compare outputs from three representative models: Kimi-K2-Instruct produces an invalid edge connection in the iteration node; GPT-5.2 generates a correct and executable workflow; GLM-4.6 outputs logically inconsistent node declarations. It reveals common failure patterns.}
    \label{fig:bad_case_study}
\end{figure*}


\subsection{Case Analysis}
We take the StudyPlanner task of Education scenario as an example for analysis. The instructions for the first two rounds are in Appendix \ref{app:instructions}.



Figure \ref{fig:bad_case_study} illustrates the second-round responses from different models. Specifically, the visual workflows generated by GPT-5.2 on the \textbf{Dify and Coze platforms} are shown in Figures \ref{dify:studyplanner2} and \ref{coze:studyplanner2} of the Appendix.
1) \textbf{Kimi-K2-Instruct} generates an invalid edge. Node \texttt{5} is an \texttt{iteration} node that internally contains a sub-workflow starting with \texttt{iteration-start}. Despite documentation explicitly restricting their relationship to inclusion only, the model erroneously connects them with an edge, rendering the workflow unexecutable.
2) \textbf{GPT-5.2} successfully resolves the case by generating a valid, executable workflow.
3) \textbf{GLM-4.6} exhibits logical inconsistencies. It utilizes the \texttt{code} and \texttt{iteration-start} nodes in the final workflow without declaring them in <node\_selection>, indicating a flawed and incoherent workflow creation logic.

\subsection{An Error-Driven Agentic Baseline}
To address the previously identified bottlenecks, we propose an error-driven agentic baseline to explore the performance upper bound of LLMs on the benchmark. Implemented via the \textbf{OpenCode} framework, this approach transitions from vanilla zero-shot prompting to a structured \textbf{SKILL-based paradigm}, explicitly incorporating task-critical guidelines, as well as rules for multi-turn interactions and variable references.


\begin{figure}[t]
    \centering
    \includegraphics[width=1.0\linewidth, trim=5 10 5 0mm, clip]{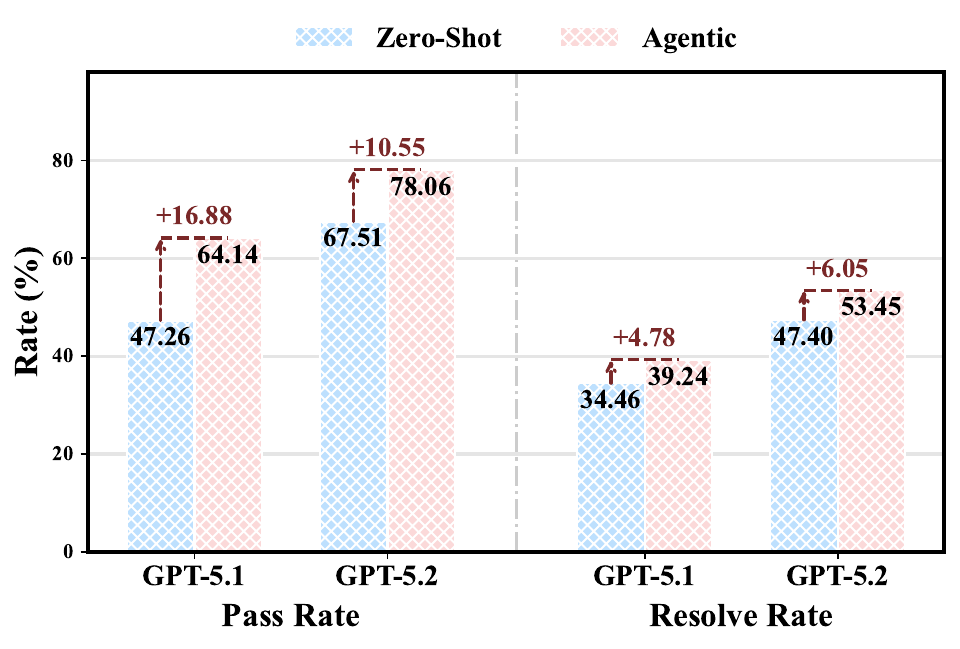}
    \vspace{-2mm}
    \caption{Performance comparison demonstrating the improvements of the proposed Agentic baseline over the Zero-Shot setting for GPT-5.1 and GPT-5.2.}
    \label{fig:baseline_performance}
\end{figure}

To mitigate context decay in subsequent interaction turns, we dynamically extract \textbf{Variable Summaries} from preceding workflow iterations as supplementary context. Crucially, we construct a robust execution loop featuring a \textbf{5-attempt retry mechanism} with comprehensive structural and semantic verification. Upon failure, the framework triggers targeted \textbf{Auto-Repair Modules} to rectify the four most prevalent errors: redundant code fences, JSON decoding failures, topological sorting violations, and node selection inconsistencies. As shown in Figure \ref{fig:baseline_performance}, the agentic framework yields substantial improvements, boosting Resolve Rate for GPT-5.1 and GPT-5.2 by absolute margins of 4.78\% and 6.05\%, respectively.

\section{Related work}

\subsection{LLM-based Agents}
LLM-based agents have demonstrated exceptional capabilities in solving complex real-world problems with the aid of general tools~\cite{schick2023toolformer,patil2024gorilla}, collaborative frameworks~\cite{metagpt,chatdev}, and extensive API repositories~\cite{toolllm,toolalpaca,xlam,wizard-code-agent}. Many prompt-based methods~\cite{wei2022chain,yao2022react,zhu2026lightthinker++} are proven effective for improving performance. However, these agent methods or frameworks mainly focus on end-to-end effects, paying little attention to the requirements and norms of the intermediate processes in solving tasks, such as planning and reasoning paths, which is not conducive to stable and reliable execution, as well as the reproduction of results across various domains~\cite{ge2024openagi,xie2024travelplanner,visualagentbench,yang2026autoskill,liang2026skillnet}.



\subsection{Workflow and Language Agent Planning}
As an intermediate state, workflows bridge the gap between high-level goals and executable steps.
By decomposing task into several executable atomic steps and organizing them into execution plans based on formal logical relationships~\cite{workflow-nets,bpmn}, the workflow achieves complete transparency of the reasoning path, effectively enhancing the interpretability and reliability of the results. 
Additionally, prior structural constraints facilitate complex task logic via established patterns~\cite{workflow-patterns,marked-DG} and robotic process automation~\cite{ivanvcic2019robotic,hofmann2020robotic}.
Although early workflows relied on labor-intensive manual design to avoid hallucinations, recent research has shifted towards automated generation using LLMs~\cite{shen2023hugginggpt,zeng2023flowmind}. This includes iterative synthesis~\cite{aflow,autoflow} and knowledge-augmented planning frameworks~\cite{ye2023proagent,knowagent,wornow2024automating,huang2024promptrpa}. Appendix \ref{app:related} provides more related work.

\section{Conclusion}
We introduce Chat2Workflow, a benchmark for evaluating LLMs' ability to generate deployable visual workflows from natural language, and show that even when augmented with an agentic framework, current models remain fragile under structural constraints and requirement changes.
Chat2Workflow can provide both a realistic testbed and a concrete target for future research on automated workflow engineering, and will help bridge the gap between language model-driven workflow design and industrial deployment.

%


\section*{Limitations}
Despite its contributions, Chat2Workflow has several limitations. First, while the high-quality dataset is manually verified , the current scale may not encompass the near-infinite variety of logic found in complex industrial business processes. Second, we simplified node interfaces to prioritize executability, which may not fully capture the intricate parameter configurations required in some real-world deployments. Finally, the current system incorporates only 20 high-frequency node types. These specific nodes were selected as they effectively cover most standard workflow scenarios. However, a wide range of valuable tools remain to be incorporated.






\bibliography{sample-base}

\clearpage
\appendix

\section{Detailed Dataset Description}
\label{data:appendix}
The description of the task scenarios is as follows:

\textbf{Research} includes 5 tasks, which focus on systematically obtaining, understanding, and integrating knowledge from unstructured data. They typically take papers, books, or research topics as input, emphasizing in-depth reading of the content, logical organization, and critical thinking skills.

\textbf{Document} includes 5 tasks, whose core value lies in transforming raw inputs such as files, images, and tables into structured results that are computable, reusable, and processable downstream. They serve as an important representative of the ``AI preprocessing layer'' in data processing.

\textbf{Enterprise} includes 4 tasks, which are targeted at real Enterprise scenarios, with the goal of automating, standardizing, and providing decision support for complex business processes through AI. Their inputs are mostly internal or business-related documents from enterprises.

\textbf{Developer} includes 3 tasks, which serve scenarios like software development, system design, and technical understanding. 
The core objective is to lower the threshold for technical understanding and development and enhance engineering efficiency.

\textbf{Education} includes 4 tasks, which revolve around the complete life cycle of ``teaching and learning'', with the goal of building a closed-loop system from learning planning, teaching content generation, assessment to feedback.

\textbf{AIGC} includes 6 tasks, which are positioned at the core of ``generating directly usable content or materials from 0 to 1'', emphasizing creative expression and multimodal output capabilities. The input is often highly abstract or brief instructions, while the output is complete, publishable, and consumable content results.

\section{Experimental Settings}
\label{app:exp}
Dify platform requires the necessary extensions to be prepared in advance. A set of common models is installed to make the workflows runnable and the results reproducible. 
Workflows in Dify are natively represented in YAML format. 
We have developed a code framework for rule-based transformation to convert the JSON format workflow generated by the language agent into valid YAML format. This ensures that workflows can be effectively generated by agents through simplified JSON representation while remaining fully compatible with the Dify execution environment. In addition, the CoT reasoning sequence is uniformly used as the output method. The node types included in the knowledge base are as follows: Start, End, LLM, Question Classifier, Code, Document Extractor, HTTP Request, If-Else, List Operator, Parameter Extractor, Template, Variable Aggregator, Iteration, Iteration-Start, Text to Speech, Text to Image, Mermaid Converter, Markdown Exporter, Google Search and Echarts. Specifically, the Question Classifier and Parameter Extractor node types are strictly designated to the Qwen-3-Max (\texttt{qwen3-max}), whereas the LLM node type is fixed to the Qwen-3-VL-Plus (\texttt{qwen3-vl-plus}). Text to Image node type are fixed to use the Z-image-Turbo (\texttt{z-image-turbo}) and Text to Speech node type are fixed to use the GPT-4o-Mini-TTS (\texttt{gpt-4o-mini-tts}). For Qwen-3-VL-Plus, temperature is 0.7, max\_tokens is 32768, other hyperparameters used during decoding are all set to default values. The system prompt and SKILL.md for workflow generation will be integrated into the GitHub repository soon. As for the Agentic Baseline method, the opencode framework we use is version 1.3.17.


%

\section{Instructions for the Studyplanner Task}
\label{app:instructions}
The instructions for the first two rounds of the Studyplanner task are as follows:

\textbf{Round 1:} Create a learning path planning workflow. The user will provide a descriptive instruction (variable \textbf{instruction}) as input. The task is to extract four core fields from the input: interested learning field, learning goals, learning preferences, and average learning duration. Subsequently, based on the extracted information, provide the user with a detailed self-study improvement plan, requiring Markdown format output. The workflow needs to output the self-study plan (variable \textbf{plan}).

\textbf{Round 2:} Modify based on the existing foundation. The workflow needs to be able to automatically generate a full set of tutorials. It requires first generating a course syllabus, and then iteratively generating knowledge points by chapter. 
The content is required to be rigorous, containing rich examples, pros and cons, and precautions. The workflow only needs to output the final tutorial integrated through template conversion (variable \textbf{tutorial}).

\section{More Related Work on Automated Workflow Evaluation}
\label{app:related}
It is crucial to evaluate the quality of the workflow created by LLM. Previous studies have tried to automate workflow evaluation in tool learning scenarios~\cite{li2023api,taskbench,xu2025comfyui} and fine-grained planning tasks~\cite{chen2024t,tooleyes,natural-plan}, using metrics to assess format validity and tool consistency. But these works still have some limitations: Firstly, Most of them remain at the stage of abstract expression of the workflow~\cite{qiao2024benchmarking}, only checking key elements without considering actual execution effects~\cite{guo2024stabletoolbench}, structural dependencies~\cite{cat-bench}, or platform-specific constraints~\cite{shen2024shortcutsbench}. They may identify obvious format errors but cannot ensure usability. Secondly, these works focus on single-round evaluation, neglecting scenarios where requirements change over time.

\begin{figure*}[!h]
\centering
\scalebox{0.85}{
\begin{tcolorbox}

You are a strict Dify Workflow evaluator. Your task is to determine whether a workflow design is valid based only on the information provided. Do not make assumptions, infer missing nodes, or introduce any external knowledge.
~\\

\# Inputs

I will provide four sections: 

1. node\_selection: A list or description of nodes selected for the workflow. It represents the nodes the designer claims will be used.

2. design\_principle: Design principles or constraints for the Dify workflow. It is used to judge overall logical consistency.

3. workflow: The JSON string representation of the Dify workflow. You must parse this JSON to extract the nodes that are actually used.

4. gt\_nodes: A list of mandatory (ground-truth) nodes. These nodes must be included to be considered valid.
~\\

\# Evaluation Rules

All rules must be satisfied for the result to be true. If any rule fails, the result must be false.
~\\

Rule 1: Ground-Truth Node Coverage

Check whether node types in gt\_nodes is a subset of node\_selection. The node type only needs to appear, and there is no need to consider the frequency of occurrence. If any node type in gt\_nodes is missing from node\_selection, immediately return false. Do not continue further checks. 
~\\

Rule 2: Consistency and Exact Node Matching

You must verify all of the following conditions:

a. Logical Consistency: node\_selection, design\_principle, and workflow must be logically consistent. The workflow structure must not violate the stated design\_principle.

b. Exact Node Set Matching (Bidirectional Constraint): Extract the actual node set from the workflow JSON. This node set must exactly match the nodes declared in node\_selection. The node type only needs to appear, and there is no need to consider the frequency of occurrence. Specifically: Nodes not declared in node\_selection must not appear in workflow. Nodes declared in node\_selection must appear in workflow.

**Note: 1. The type of the "Template" node is "template-transform". 
2. Node matching is case-insensitive and ignores hyphens.
3. Treat aliases as identical (e.g., "tts" = "Text to Speech", "text2image" = "Text to Image").**
~\\

\# Final Decision Logic

Return true only if Rule 1 and Rule 2 are fully satisfied. If any single condition fails, return false. There is no partial credit or “mostly correct” outcome.
~\\

\# Output Format

Your response must follow this exact format and nothing else:

<reason>

[Explain the evaluation reasoning, stating which rules were satisfied or violated]

</reason>

<result>

[true or false]

</result>
\end{tcolorbox}
}
\caption{Prompt for evaluating pass rate.}
\label{prompt:pass rate}
\end{figure*}

\begin{figure*}[!h]
\centering
\scalebox{0.85}{
\begin{tcolorbox}

You are an expert evaluator of workflow execution quality. Your task is to judge whether the execution result of a workflow satisfies the requirements of the current round instruction, based on all available information.
~\\

\# Inputs You Will Be Given

You will be provided with the following four parts:

1. queries: A list of historical instructions for creating or modifying the workflow.

\quad - Each round’s instruction is based on the previous ones and may add, modify, or refine workflow behavior.

\quad - The latest instruction represents the requirements for the current evaluation round.

\quad - Ignore the input and output requirements for the file in the instruction of query. The file part has been evaluated separately.

2. input: The non-file input variables used in the workflow execution of the current round.

\quad - This field may be empty.

3. output: The non-file output variables produced by the workflow execution of the current round.

4. reference\_answer: A reference answer for the expected result.

\quad - It may contain minor omissions.

\quad - Its format may not strictly match the required output format.

\quad - However, its content is considered correct.

\quad - This field may be empty.
~\\

\# Evaluation Instructions

Please evaluate whether the workflow execution meets the requirements of the current round instruction, following these principles:

1. Instruction Alignment

\quad - Identify the requirements implied or explicitly stated in the latest instruction within queries.

\quad - Consider earlier instructions only insofar as they remain valid and are not overridden.

2. Information Utilization

\quad - Use all available information (queries, input, output, reference\_answer) for judgment.

\quad - Both of input and reference\_answer may be empty; this alone does not imply failure.

3. Output Correctness Criteria

\quad - If output exists, judge whether its content and format satisfy the current instruction, except for the requirements related to the files.

\quad - If reference\_answer exists, use it as a semantic reference rather than a strict template.

4. Handling Missing or Empty Fields

\quad - If outputs clearly contradict the instruction, the result should be false, except for the requirements related to the files.

\quad - If reference\_answer is empty, rely on instruction compliance and logical consistency only.

5. Final Judgment Rule

\quad - If the workflow execution reasonably fulfills the intent and requirements of the current round instruction (except for the requirements related to the files), return true.

\quad - Otherwise, return false.

**Note: This rule takes precedence over all the above —— The file variable part in the instruction requirements, whether as input or output, including images, documents, and audio will not be provided in this evaluation. So it cannot be used as the basis for a judgment of 'False'!!! For example, if the instruction or query requires the output of a PDF document but the output does not include the document, this point should be ignored at this time to check whether other aspects meet the requirements. **
~\\

\# Output Format

Your response must follow this exact format and nothing else:

<reason>

[Explain the evaluation reasoning, stating which rules were satisfied or violated]

</reason>

<result>

[true or false]

</result>

\end{tcolorbox}
}
\caption{Prompt for evaluating resolve rate.}
\label{prompt:resolve rate}
\end{figure*}


\begin{figure*}[t]
    \centering
    \includegraphics[width=1.0\linewidth]{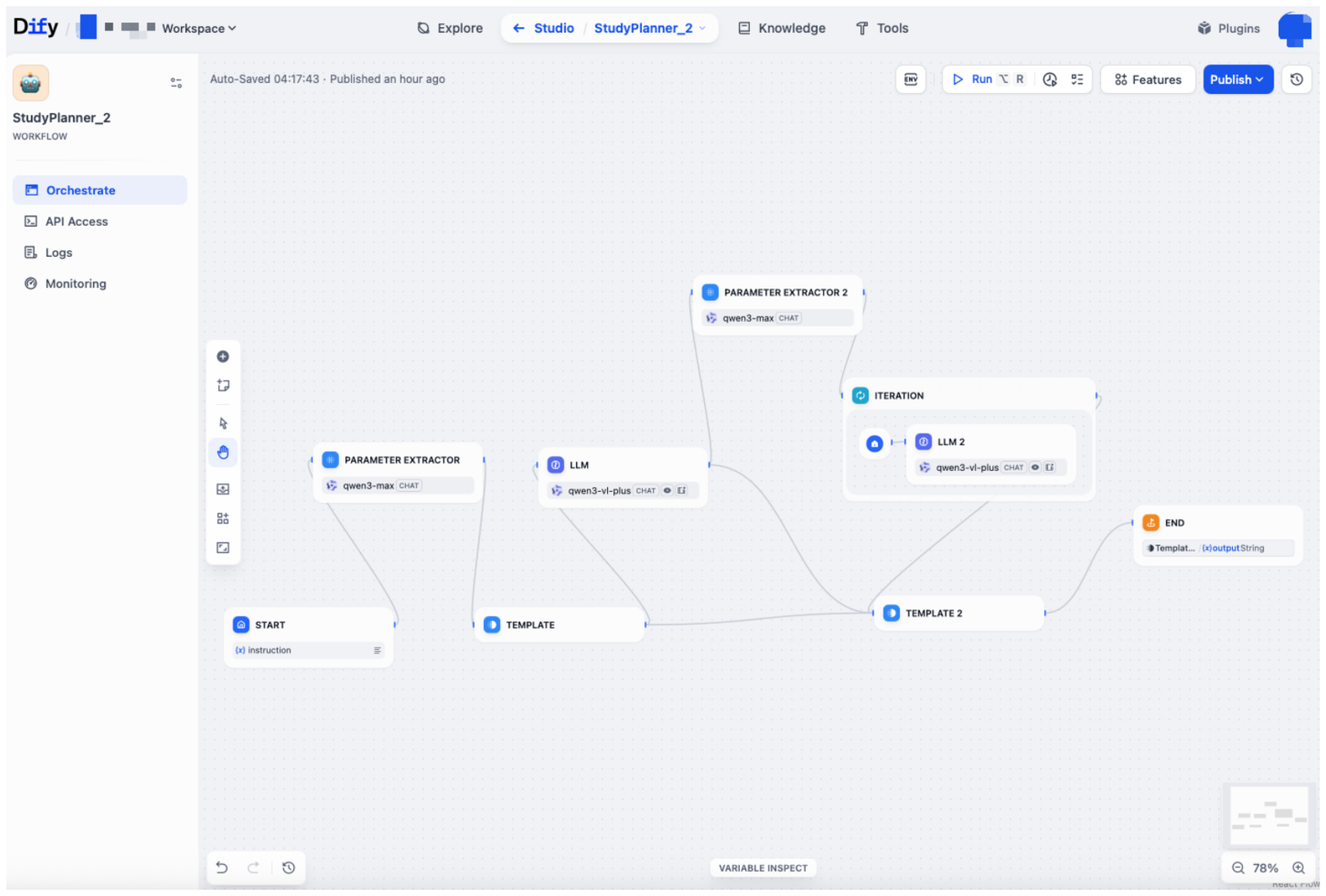}
    \caption{The Dify workflow generated by GPT-5.2 in the second round of the Studyplanner task.}
    \label{dify:studyplanner2}
\end{figure*}

\begin{figure*}[t]
    \centering
    \includegraphics[width=1.0\linewidth]{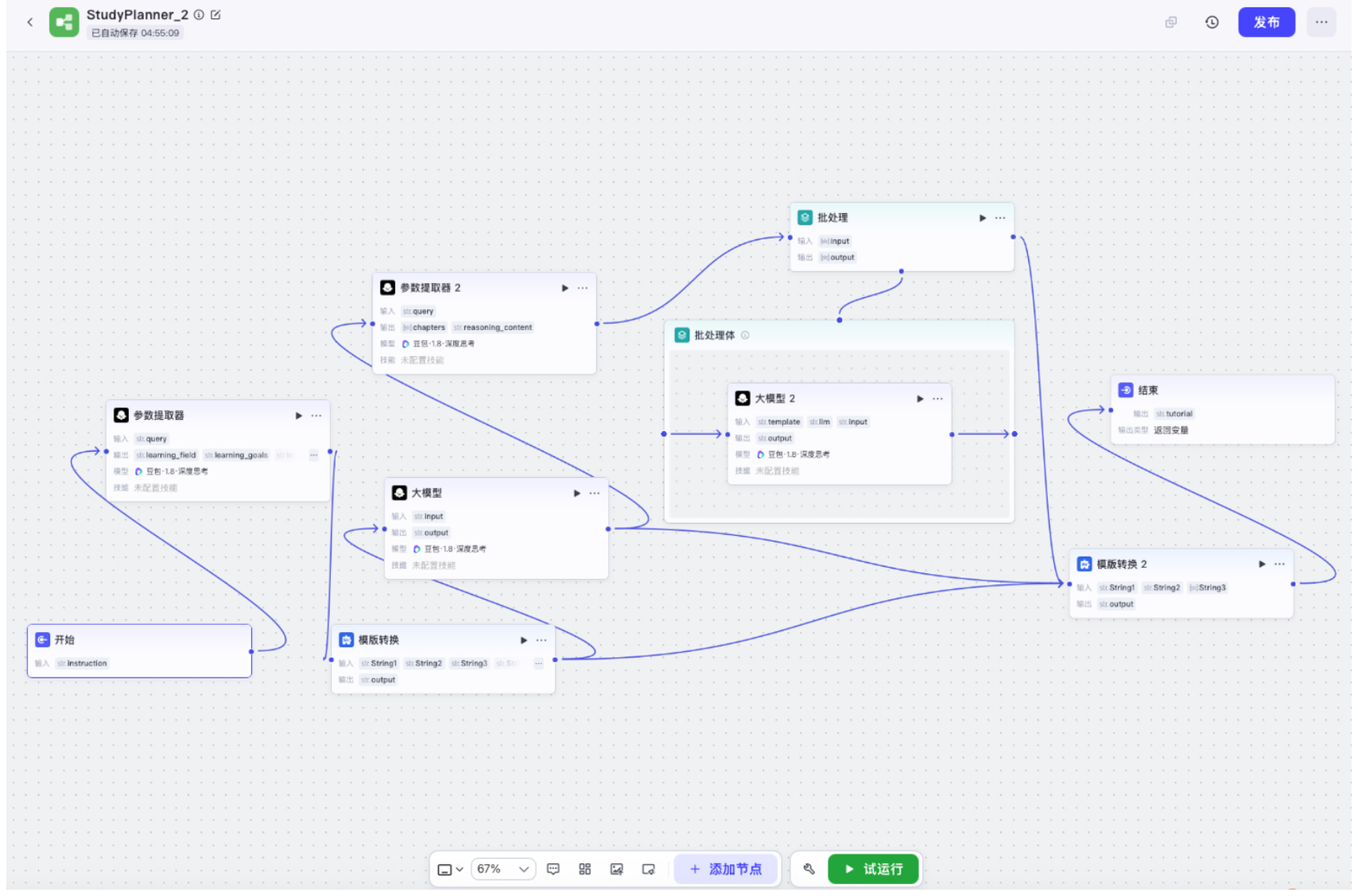}
    \caption{The Coze workflow generated by GPT-5.2 in the second round of the Studyplanner task.}
    \label{coze:studyplanner2}
\end{figure*}


\end{document}